\documentclass[final,5p,times,twocolumn]{elsarticle}

\usepackage{framed,multirow}
\usepackage{graphicx}
\usepackage{amssymb}
\usepackage{amsthm}

\usepackage{url}
\usepackage{xcolor}
\usepackage{hyperref}

\usepackage{latexsym}
\usepackage{cite}
\usepackage{amsmath,amssymb,amsfonts}
\usepackage{algorithmic}
\usepackage{graphicx}
\usepackage{textcomp}
\usepackage{caption}
\usepackage{booktabs}
\usepackage{array}
\usepackage{bbding}
\usepackage{pifont}
\usepackage{subcaption}
\usepackage{url}
\usepackage{hyperref}
\usepackage{xcolor}

\begin{document}

\begin{frontmatter}


\title{TiAVox: Time-aware Attenuation Voxels for Sparse-view 4D DSA Reconstruction}




\author[1]{Zhenghong~Zhou\corref{cor1}}
\author[2]{Huangxuan~Zhao\corref{cor1}}
\cortext[cor1]{Equal~contribution.   $^{\dagger}$ Corresponding author.}
\author[1]{Jiemin~Fang}
\author[2]{Dongqiao~Xiang}
\author[2]{Lei~Chen}
\author[2]{Lingxia~Wu}
\author[2]{Feihong~Wu}
\author[1]{Wenyu~Liu}
\author[2]{Chuansheng~Zheng$^{\dagger,}$}
\ead{hqzcsxh@sina.com}
\author[1]{Xinggang~Wang$^{\dagger,}$}
\ead{xgwang@hust.edu.cn}

\address[1]{School of Electronic Information and Communications, Huazhong University of Science and Technology, Wuhan, China}
\address[2]{Department of Radiology, Union Hospital, Tongji Medical College, Huazhong University of Science and Technology, Wuhan, China}


\begin{abstract}
Four-dimensional Digital Subtraction Angiography (4D DSA) plays a critical role in the diagnosis of many medical diseases, such as Arteriovenous Malformations (AVM) and Arteriovenous Fistulas (AVF). However, radiocontrast flow and intricate vessels makes it difficult to reconstruct high quality 4D DSA image, traditional method like Feldkamp-Davis-Kress (FDK) demands numerous views to reconstruct, thereby implying a significant radiation dose. And other medical neural radiance field methods such as NAF are not applicable in situations involving contrast agent flow. To address this radiocontrast flow and high radiation issue, we propose a Time-aware Attenuation Voxel (TiAVox) approach for sparse-view 4D DSA reconstruction, which paves the way for high-quality 4D imaging. Additionally, 2D and 3D DSA imaging results can be generated from the reconstructed 4D DSA images. TiAVox introduces 4D attenuation voxel grids, which reflect attenuation properties from both spatial and temporal dimensions. It is optimized by minimizing discrepancies between the rendered images and sparse 2D DSA images. Without any neural network involved, TiAVox enjoys specific physical interpretability. The parameters of each learnable voxel represent the attenuation coefficients. We validated the TiAVox approach on both clinical and simulated datasets, achieving a 31.23 Peak Signal-to-Noise Ratio (PSNR) for novel view synthesis using only 30 views on the clinically sourced dataset, whereas traditional Feldkamp-Davis-Kress methods required 133 views. Similarly, with merely 10 views from the synthetic dataset, TiAVox yielded a PSNR of 34.32 for novel view synthesis and 41.40 for 3D reconstruction. We also executed ablation studies to corroborate the essential components of TiAVox. The code will be publicly available.
\end{abstract}


\end{frontmatter}


\section{Introduction}
\label{sec:introduction}
4D DSA images deliver superior insights into blood flow and vascular structures compared to 2D or 3D DSA images and have been commercialized as a prominent technique for interventional neuroradiology. Offering three-dimensional, time-resolved images, 4D DSA permits observation of the contrast agent's bolus traversing the vascular system from any desired perspective at any given instance. This diagnostic tool is paramount in evaluating conditions such as AVM, AVF, intracranial aneurysms, and occlusive atherosclerotic diseases \citep{4ddsa, sandoval20164d, lang20174d, sandoval2015comparison}. At present, the reconstruction of 4D DSA images employs the classical and ubiquitously used FDK reconstruction algorithm \citep{feldkampPracticalConebeamAlgorithm1984}. However, this procedure requires hundreds of 2D DSA images, captured from a range of viewpoints and times, which results in substantial radiation exposure posing potential harm to patients.

\begin{table}[tbp]
\centering
\renewcommand{\arraystretch}{1.3}
\caption{Compare with other methods. FDK is the method utilized in commercial medical imaging devices for reconstruction. CNN represents Convolutional Neural Network-based methods used for sparse CT reconstruction. NeRF-in-Med refers to Neural Radiance Fields (NeRF) methods applied in medical reconstruction.}
\label{tab:3method}
\centering
\setlength{\tabcolsep}{1.0mm}
\scalebox{0.8}
{
\begin{tabular}{ccccc}
\toprule
 & FDK & CNN & NeRF-in-Med & TiAVox\\
\midrule
Sparse view &  & \Checkmark & \Checkmark & \Checkmark    \\
Interpretable parameters & \Checkmark &  &  & \Checkmark    \\
Suitable for radiocontrast flow & \Checkmark &  &  & \Checkmark     \\
No need for 3D label & \Checkmark &  & \Checkmark & \Checkmark    \\
Space-time 4D result & \Checkmark &  &  & \Checkmark    \\

\bottomrule
\end{tabular}}

\end{table}

The goal is to lower the radiation dose from 4D DSA imaging by decreasing the number of 2D images collected. The emergence of deep learning methods has prompted various studies on sparse-view 3D computed tomography (CT)/DSA reconstruction using convolutional neural networks (CNN) \citep{jinDeepConvolutionalNeural2017, hanFramingUNetDeep2018, zhuImageReconstructionDomaintransform2018, zhangSparseViewCTReconstruction2018, zhao2022self, shan20183}. However, the nature of CNNs as non-interpretable black box models, with an unclear feature extraction process, presents obstacles for practical applications. Moreover, these methods are not directly applicable to 4D DSA reconstruction for the following reasons: (a) their primary suitability is for static CT reconstruction, not accounting for the flow of the radiocontrast agent. Consequently, inconsistent vessel information, observed at different views and times, could degrade the reconstruction performance; (b) CNN-based 3D CT reconstruction methods necessitate precise 3D labels for training. Unfortunately, the 3D DSA labels, obtained via the gold standard method (FDK), require intricate post-processing with deformation. This inevitably leads to misalignment with the input 2D DSA image, rendering CNN training on clinically collected datasets challenging.

\begin{figure*}[tbp]
\centering
\includegraphics[width=\linewidth]{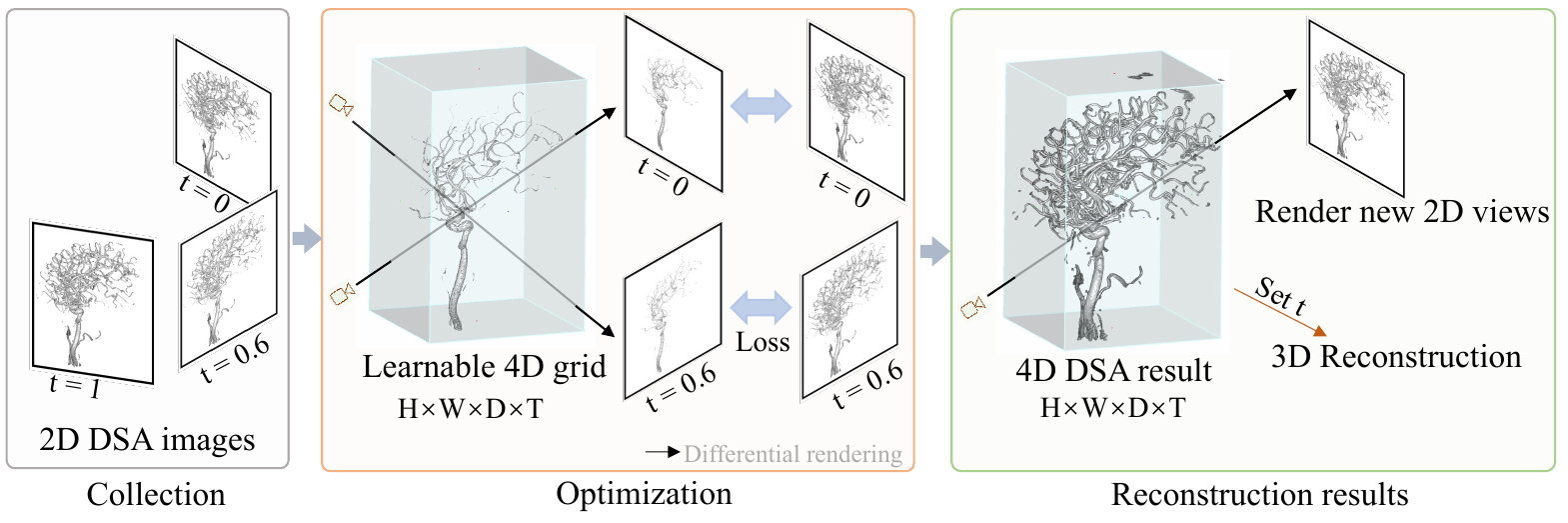}
\caption{\textbf{Pipeline of TiAVox}, TiAVox is a learnable 4D voxel grid, that can generate any views by differential rendering and is optimized through a set of 2D DSA \{projection, pose, time\} from scratch. When optimization is finished, TiAVox could be directly as the result of 4D DSA, and further get 3D DSA at a certain moment and 2D DSA through rendering.}
\label{fig:pipelineofdsarf}
\end{figure*}

In recent years, Neural Radiance Field (NeRF) methods \citep{mildenhallNeRFRepresentingScenes2020a} have been highly successful in 3D reconstruction tasks and novel view synthesis using natural images. Their success primarily relies on two aspects: (a) the employment of differentiable rendering for efficient supervision and (b) the utilization of deep learning models to represent specific scene information. Numerous subsequent studies have improved NeRF, encompassing dynamic scenes \citep{pumarola2021d, parkHyperNeRFHigherDimensionalRepresentation2021, TiNeuVox, tensor4d, TemporalInterpolation, nerfplayer,dynamicsurface}, sparse-view scenarios \citep{niemeyerRegNeRFRegularizingNeural, ViPNeRF, consistentnerf}, and modifications for explicit representation to expedite the process \citep{yuPlenoxelsRadianceFields2021a, sunDirectVoxelGrid2021, tensorf, trimiprf,voxurf}. NeRF employs the same rendering equation and camera model as DSA/CT imaging, and it exhibits exceptional potential in scalability and superior reconstruction results. It is worth noting that, in the case of monocular dynamic scene reconstruction of natural images, many methods employ explicit representations along with deformation fields to model dynamic scenes\citep{TiNeuVox, tensor4d}. Directly utilizing explicit structures without motion fields is hard due to the limited information available in monocular videos compared to the entire dynamic scene. We demonstrate, in the context of monocular 4D DSA reconstruction, that a purely explicit 4D grid structure, employed by TiAVox, is capable of achieving high-quality reconstruction results even in the absence of regularization.

In fact, there are some recent works that apply NeRF to medical situations, such as CT reconstruction \citep{corona2022mednerf, ruckert2022neat, zha2022naf, shen2022nerp, snaf, wu2022self, zang2021intratomo, lin2023learning} and deformable tissues reconstruction \citep{yang2023neural, wang2022neural}, etc. Medical surface reconstruction also has some work like Going Off-Grid \citep{alblas2022going}. Nevertheless, these studies commonly neglect the radiocontrast flow, resulting in frequent reconstruction failures in such scenarios. For instance, in Figure 5, the results from NAF~\citep{zha2022naf} exhibit partial blurring in certain blood vessels. In contrast to static medical reconstruction, the dynamics of radiocontrast flow induce temporal changes in the vessel's internal signals. A recent work~\citep{reed2021dynamic} uses implicit neural representations for 4D-CT reconstruction. Due to the adoption of implicit representation and the deformation field, this has resulted in lower optimization efficiency and accuracy. Furthermore, unlike scenarios involving rigid body or tissue movement, in DSA imaging, the blood vessels remain stationary, while the radiocontrast agent's signal within the vessels undergoes alterations, posing a unique dynamic medical reconstruction challenge.

To model the radiocontrast flow and facilitate sparse-view 4D DSA image reconstruction, we propose the Time-aware Attenuation Voxels (TiAVox) method. A 4D grid is constructed with both the 3D spatial dimension and the 1D temporal dimension, the latter of which allows for modeling the radiocontrast flow. TiAVox works under a self-supervised paradigm, solely supervised by sparse-view 2D DSA images, eliminating the need for 3D supervision. Unlike typical neural network-based architectures which often lack interpretability, TiAVox adopts an explicit structure where each parameter carries a specific physical interpretation as an attenuation coefficient. In addition, this explicit architecture significantly accelerates the reconstruction speed compared to implicit architecture. 

We validated TiAVox in clinically collected and simulated data from Wuhan Union Hospital. From the predicted 4D DSA results, we render 2D images from novel views and 3D results by setting a specific time. These 2D and 3D results were used to evaluate TiAVox's performance. The clinically collected DSA images, influenced by factors like the flow of the radiocontrast agent, noise, and minor discrepancies in camera pose, were accurately reconstructed by TiAVox with only 30 views, achieving an over 31 2D PSNR value for novel view synthesis. For the simulated datasets, TiAVox efficiently reconstructs high-quality results with mere 10 views — only 1/13 of the views required by the gold standard method (FDK). This achieves a high 3D PSNR of 41.4 and a 2D PSNR of 34.32.

Our contributions can be summarized as:
\begin{itemize}
\item To the best of our knowledge, we propose the first NeRF-based 4D DSA reconstruction method, only requiring a few 2D views as input/supervision, which significantly reduces the radiation dose, and is suitable for 4D DSA imaging with radiocontrast flow.
\item TiAVox is a transparent and explainable 4D DSA reconstruction method, containing 4D grids for both 3D spatial and 1D temporal modeling, and does not contain any black-box neural networks.
\item TiAVox can produce high-quality 4D/3D/2D DSA imaging results on both clinically collected and simulated DSA datasets, outperforming other strong NeRF-based medical imaging methods, such as NAF~\citep{zha2022naf} and DirectVoxGo~\citep{sunDirectVoxelGrid2021}, with fast convergence speed.
\end{itemize}

\section{Method}
\label{sec:method}

In this section, we begin with outlining the overall framework of TiAVox, including the task definition and the method pipeline. Next, we affirm the suitability of NeRF for CT/DSA imaging. Subsequently, we illuminate the process of representing 4D DSA scenes using TiAVox. Finally, we detail the optimization process inherent to TiAVox.

\subsection{The overall framework of TiAVox}
\label{sec:method_framework}
We focus on sparse views 4D DSA reconstruction task. Given a sequence of sparse 2D DSA projections $P$ paired with time $t$ and denoted as ${\{P_1,t_1;P_2,t_2...P_M,t_M\}}$, along with their corresponding camera parameters ${\{C_1,C_2...C_M\}}$, the goal is to derive the 4D DSA output $O$. This problem pivots on learning a mapping function, mathematically expressed as $\mathcal{M}:(\boldsymbol{x}, t)\rightarrow(\sigma)$, that associates coordinates with attenuation coefficients/densities $(\sigma)$. When $\mathcal{M}$ is suitably trained, the 4D DSA output $O$ is directly procured by traversing all voxel coordinates and time, then applying the model $\mathcal{M}$ for prediction. Additionally, by setting a specific time, the 3D DSA result is extracted, and by rendering in the given camera pose and time, we achieve the 2D DSA result of novel view synthesis.

We propose a fully explicit, time-aware attenuation voxels (TiAVox) method to represent the mapping function $\mathcal{M}$. TiAVox functions as a learnable 4D voxel grid, wherein each voxel denotes the attenuation coefficient pertinent to its corresponding position and time. For any given point $(\boldsymbol{x}, t )$, TiAVox can procure the corresponding attenuation coefficients through interpolation in both spatial and temporal domains.

Fig.~\ref{fig:pipelineofdsarf} illustrates the pipeline of TiAVox. Initially, TiAVox is set to a minimal value and rays are generated based on the camera parameters. Subsequently, points along the ray interact with TiAVox, retrieving attenuation coefficients to compute 2D rendering projections. The discrepancy between these projections and the 2D DSA images serves as a basis to optimize TiAVox through backpropagation.

\begin{figure}[tbp]
\centering
\includegraphics[width=0.8\linewidth]{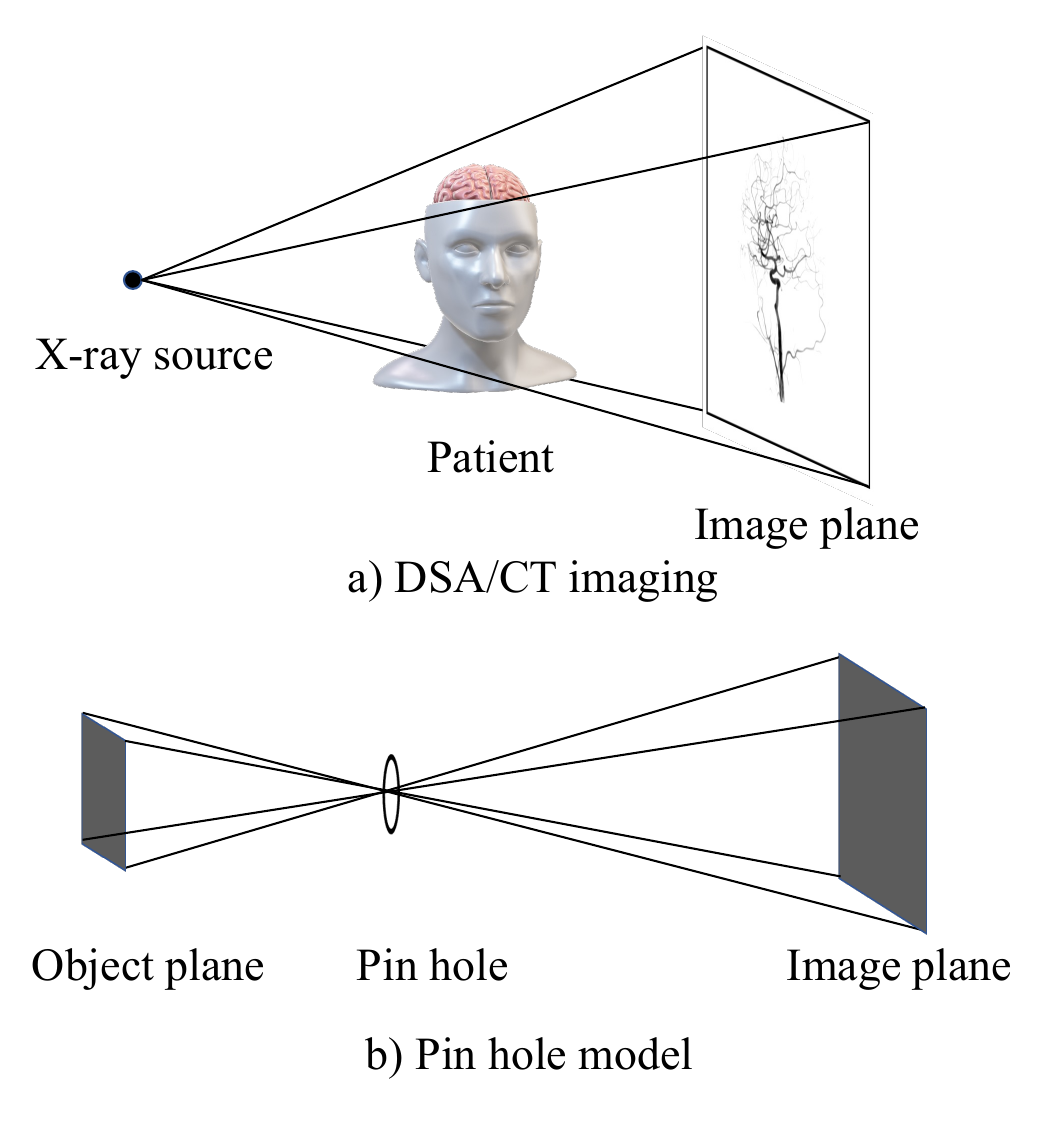}
\caption{DSA VS NeRF, DSA/CT imaging and NeRF pinhole model are aligned in render equation and camera model.}
\label{fig:DSAVSNeRF}
\end{figure}

\subsection{Formulation of the imaging process}
The formation of the imaging process consists of two main aspects: the rendering equation and the imaging model. It is worth noting that the pinhole camera model and rendering equation used in NeRF \citep{mildenhallNeRFRepresentingScenes2020a} can be well aligned to the CT reconstruction.

\textbf{Render equation}: According to the Beer–Lambert law, the intensity of the X-rays decays exponentially through the medium and is expressed mathematically as:

\begin{equation}
I=I_0\exp{\left(-\sum_{i=1}^{N}{\sigma_i\delta_i}\right)\ },
\label{equ:render}
\end{equation}
where ${I_0}$ is initial intensity of X-ray, ${\sigma_i}$ denotes the attenuation coefficient at $i$ location, ${\sigma_i}$ represents the distance between two points.

\textbf{Pinhole camera model}: As shown in Fig.~\ref{fig:DSAVSNeRF}, the pinhole imaging model used in NeRF model can be equivalent to the imaging model used in DSA equipment. The pinhole in (b) can be seen as X-ray source in (a), the object plane in (b) corresponds to the patient in (a), and the image plane in (b) corresponds to the image plane in (a). This makes the NeRF-like method easy to use in CT/DSA reconstruction. 

\begin{figure*}[tbp]
\centering
\includegraphics[width=\linewidth]{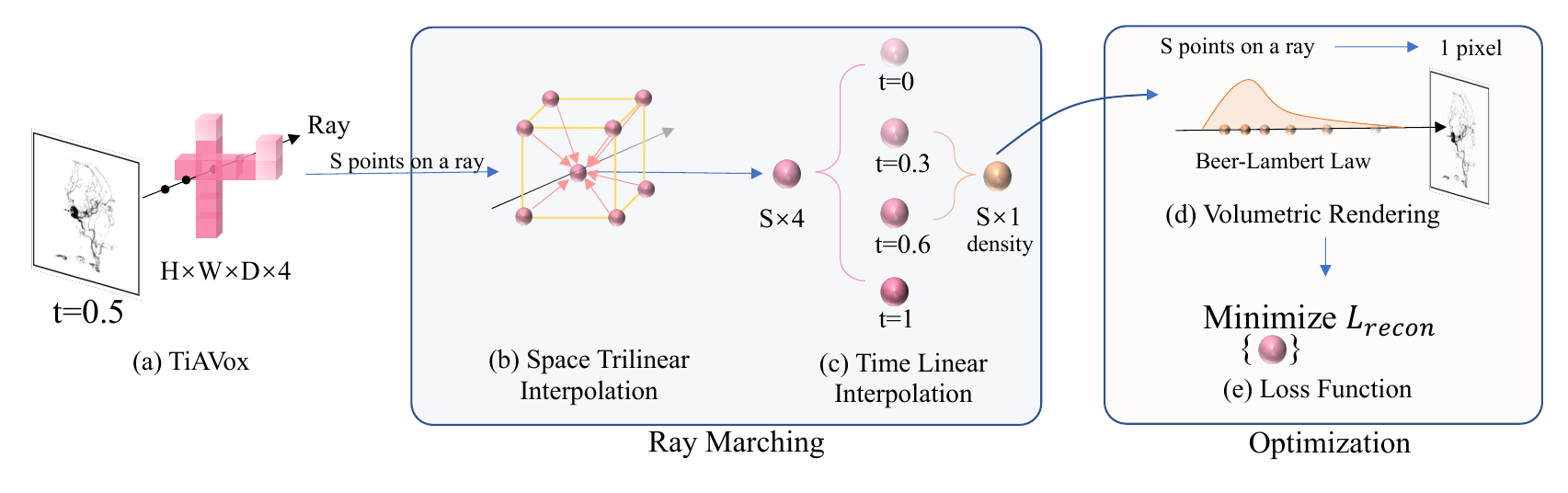}
\caption{\textbf{Optimization of TiAVox}, including ray marching and optimization part, in ray marching, points on a ray could query attenuation coefficient by spatial trilinear interpolation and temporal linear interpolation; in optimization part, attenuation coefficient on a ray composite a pixel by volumetric differential rendering and compute loss with pixel in 2D DSA image.}
\label{fig:Optimization}
\end{figure*}

\subsection{Time-aware attenuation voxels}

\textbf{Ray sample}. Each pixel value in a projection image is the result of X-ray attenuation as it passes through a medium. In ray sample, rays $\{r_1,r_2...\}$ are generated using camera parameters ${\{C_1,C_2...C_M\}}$, and points $\{(\boldsymbol{x_1}, t_1), (\boldsymbol{x_2}, t_3)...\}$ are sampled along these rays. Noting that a single ray samples multiple points. Similar to DirectVoxGo \citep{sunDirectVoxelGrid2021}, we perform point sampling at regular intervals by setting the sampling interval parameter. And we also exclude points that do not intersect with the known free space.

\textbf{TiAVox} is a 4D voxel grid devoid of any neural network, directly represents the attenuation coefficients of any 3D points at any given point in time. Contrary to other methodologies that simulate object motion within a dynamic scene via a deformation network \citep{TiNeuVox, pumarola2021d, yang2023neural}, we model and reconstruct the entirety of the spatiotemporal information directly. This method is highly adaptable and uniquely suited to capture vascular information alterations during the flow of contrast agents, given that the vessels themselves remain static. Moreover, the adoption of an entirely explicit architecture yields enhanced interpretability and accelerates computational speed.

Given any points at any time, the attenuation coefficient can be directly queried by spatial trilinear interpolation and temporal linear interpolation of TiAVox, as shown in Fig.~\ref{fig:Optimization}: 

\begin{equation}
\operatorname{interp_{spatial}}(\boldsymbol{x}, \boldsymbol{G}):\left(\mathbb{R}^3, \mathbb{R}^{N_t \times N_h \times N_w \times N_d}\right) \rightarrow \mathbb{R}^{N_t},
\end{equation}

\begin{equation}
\operatorname{interp_{temporal}}(t, \boldsymbol{G_x}):\left(\mathbb{R}^1, \mathbb{R}^{N_t}\right) \rightarrow \mathbb{R}^1,
\end{equation}
where $\boldsymbol{x}$ represents the spatial location of the queried point. $\boldsymbol{G}$ is the 4D voxel grid of TiAVox with size $N_t \times N_h \times N_w \times N_d$, where corresponding to time, height, width and depth, respectively. $t$ is the time of queried point. $\boldsymbol{G_x}$ is the result of ${interp_{spatial}}(\boldsymbol{x}, \boldsymbol{G})$. The final result of ${interp_{temporal}}(t, \boldsymbol{G_x})$ is attenuation coefficient ${\sigma}$ of the queried point $(\boldsymbol{x}, t)$.

Time interpolation enables TiAVox to learn the  attenuation coefficient that changes dynamically with time. $N_t$ can be viewed as temporal resolution. In Fig.~\ref{fig:Optimization}, $N_t$ = 4, each spatial point has 4 values referring to the vessel density at $t$ = 0, 0.3, 0.6, 1. 

\subsection{Model optimization}

\textbf{Optimization of TiAVox}. After the attenuation coefficient of queried points is acquired, the final value of an X-ray hitting the receiving plate can be calculated according to the rendering Eq.~\eqref{equ:render}, and the TiAVox is optimized by back-propagating the difference between the rendered result and the 2D DSA image. The loss function is:

\begin{equation}
\mathcal{L}_{\text {recon}}(\boldsymbol{G})=\frac{1}{|\mathcal{B}|} \sum_{r \in \mathcal{B}}\|\hat{I}(\boldsymbol{r})-I(\boldsymbol{r})\|_2^2,
\end{equation}
where $\mathcal{B}$ represents the set of rays in a sampled mini-batch. $\hat{I}$ and $I$ are prediction and ground truth projections for ray $r$ respectively.

\textbf{Progressive temporal resolution scaling}. Inspired by NSVF \citep{liu2020neural} and DirectVoxGo \citep{sunDirectVoxelGrid2021}, we optimize the temporal resolution of TiAVox from coarse to fine. Assuming an initial temporal resolution of 4, when reaching the specified iterations, a new value is linearly interpolated between every two adjacent values, resulting in an expanded temporal resolution of 7. For instance, a value of $t$ = 0.15 is inserted between $t$=0 and $t$ = 0.3.

\section{Experiment}

\begin{figure}[tbp]
\begin{center}
\includegraphics[width=\linewidth]{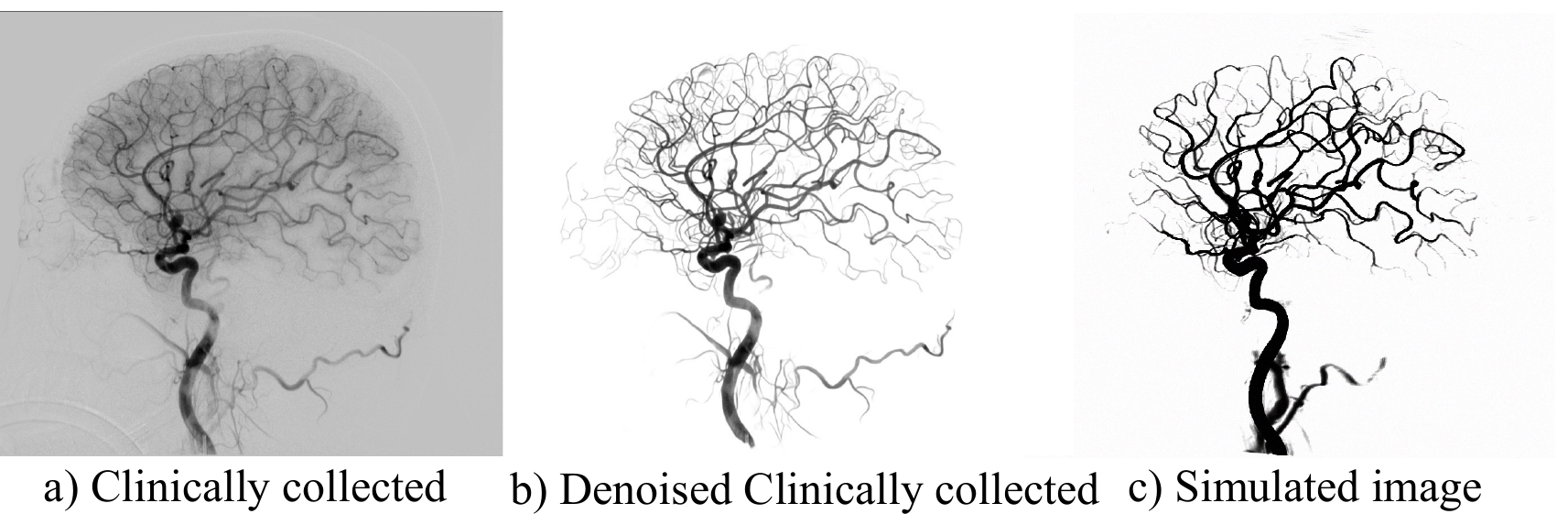}
\end{center}
\caption{Contrast of clinically collected 2D DSA image, denoised clinically collected 2D DSA image, and simulated 2D DSA image.}
\label{fig:denoise}
\end{figure}


\subsection{Data and experiment setting}
The datasets in this study of different eight patients was retrospectively collected from Wuhan Union Hospital. The patients being treated at the participating facilities were approached in accordance with the institutional review board-approved protocols, and they provided their informed consent. For each patient data, a total of 133 2D projection images were recorded, and the images were transferred to the workplace workstation for 3D reconstruction (gold standard), we refer to this result as 133-view FDK 3D result in the paper. 

Relevant information was obtained from the patient's DICOM file, including 2D pixel array, main angle, secondary angle, the distance from the transmitter to the receiver and pixel spacing. Since the distance from the patient to the transmitter cannot be acquired straightly, we estimated half of the distance from the transmitter to the receiver. This information is transformed into camera poses used for reconstruction.

Since we can not access 4D DSA images, and only have 133-view 2D DSA images as well as 133-view FDK 3D results, we present the results of TiAVox in tasks of \textbf{2D novel view synthesis} and \textbf{3D reconstruction}. As delineated in Sec.~\ref{sec:method_framework}, TiAVox exhibits the capacity to generate 2D and 3D DSA images. Furthermore, we display results under two experimental conditions: reconstruction from \textbf{clinically collected projections} and from \textbf{simulated projections} (digitally generated projections) with precise camera positioning. In the case of simulated projections, we directly apply identical volume rendering to the 133-view FDK 3D result, preserving the camera parameters of each rendering. We introduce Gaussian noise to each projection, bearing a standard deviation equivalent to 0.02 times the maximum intensity. Despite the absence of contrast agent effects in this scenario, it serves as a valid benchmark to evaluate the effectiveness of the method. The visualization of datasets are shown in Fig.~\ref{fig:denoise}.

\begin{figure*}[!t]
\centering
\includegraphics[width=450pt]{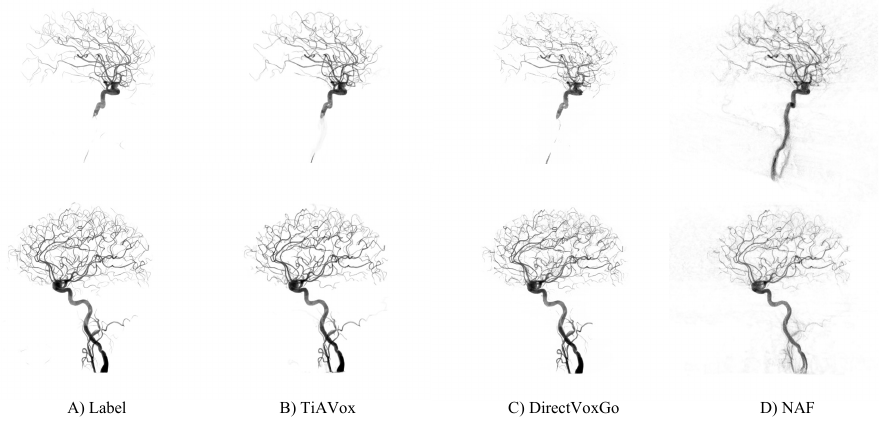}
\caption{Novel view synthesis result of clinically collected images (30 views)}
\label{fig:realresult}
\end{figure*}

\begin{table*}[!t]
    \centering
    \renewcommand{\arraystretch}{1.2}
    \caption{Novel view synthesis 2D PSNR and SSIM results from clinically collected images over views (8 patients)}
    \label{tab:resultoverview}

    \scalebox{0.9}{
    \begin{tabular}{>{\centering\arraybackslash}p{4.5cm} *{12}{>{\centering\arraybackslash}p{0.6cm}}>{\centering\arraybackslash}p{1.5cm}}
        \toprule
        \multicolumn{1}{c}{Views} & \multicolumn{2}{c}{4 views} & \multicolumn{2}{c}{10 views} & \multicolumn{2}{c}{20 views} & \multicolumn{2}{c}{30 views} & \multicolumn{2}{c}{40 views} & \multicolumn{2}{c}{50 views} & \multicolumn{1}{c}{Training time} \\
         & PSNR$\uparrow$ & SSIM$\uparrow$ &  PSNR$\uparrow$ & SSIM$\uparrow$ &  PSNR$\uparrow$ & SSIM$\uparrow$ & PSNR$\uparrow$ & SSIM$\uparrow$ & PSNR$\uparrow$ & SSIM$\uparrow$ & PSNR$\uparrow$ & SSIM$\uparrow$ & (30 views) \\
        \midrule
        NAF \citep{zha2022naf} & 24.30 & 0.83 & 26.06 & 0.85 & 26.78 & 0.88& 26.79 & 0.90 & 26.89 & 0.90 & 27.07 & 0.91 & \textgreater 30 min \\
        DirectVoxGo \citep{sunDirectVoxelGrid2021} & 22.48 & 0.87 & 24.85 & 0.91 & 26.86 & 0.92 & 28.23 & 0.93 & 29.18 & 0.94 & 29.82 & 0.94 & 21.1 min \\
        TiAVox (ours) & \textbf{24.99} & \textbf{0.90} & \textbf{28.57} & \textbf{0.94} & \textbf{30.46} & \textbf{0.95} & \textbf{31.23} & \textbf{0.96} & \textbf{31.51} & \textbf{0.96} & \textbf{31.94} & \textbf{0.96} & \textbf{4.9 min}\\

        \bottomrule
    \end{tabular}}

\end{table*}

Simulated projections merit consideration for two primary reasons. Firstly, due to the discrepancy in the reconstruction methods, the 133-view FDK 3D results (the gold standard) and the reconstructed results do not align precisely. Consequently, the 3D outcomes derived from our method cannot be directly compared with the gold standard to yield quantitative indicators such as PSNR. In contrast, simulated data can be evaluated using PSNR. Secondly, clinically collected images often contain inaccurate camera parameters, including the distance from the patient to the X-ray transmitter. These inaccuracies can influence the reconstruction result. Therefore, we aim to demonstrate the reconstruction performance of TiAVox in the absence of pose error and non-vascular noise.

\subsection{Implementation details}

Clinically collected DSA images often contain various sources of noise, such as bone signals, imaging artifacts, and other disturbances. These can considerably impact the quality of novel view synthesis and 3D reconstruction. It's noteworthy that the desired result does not necessitate the reconstruction of signals outside the blood vessels. To address this, we first aim to minimize the noise in the input DSA images. For this purpose, we employ a U-Net \citep{ronnebergerUNetConvolutionalNetworks2015}. As illustrated in Fig.~\ref{fig:denoise}, noise reduction effectively eliminates interfering signals, enabling TiAVox to deliver better reconstruction results.

In the implementation of TiAVox, we implement our model based on PyTorch \citep{pytorch}, reference the code of DirectVoxGo \citep{sunDirectVoxelGrid2021}, and use progressive scaling, post-activated density voxel grid and free space skipping strategy. All our experiments are completed on a single RTX 3090 GPU or NVIDIA Tesla V100 GPU. We utilize the Adam optimizer \citep{kingmaAdamMethodStochastic2017} with a batch size of 8,192 rays, optimizing for 20k iterations. Learning rate is 0.1 and exponential learning rate decay strategy is employed. The resolution of training and validation images is $1024 \times 1024$.

For clinically collected images, we set the final spatial resolution as $320^3$ and temporal resolution as 4, which progressively increases to 7. For simulated images, we set the final spatial resolution as $470^3$ and the temporal resolution as 1.

\subsection{Reconstruction from clinically collected images}

We perform reconstruction using 4-50 views (4, 10, 20, 30, 40, 50) respectively, and show the results of 3D reconstruction and novel view synthesis. DirectVoxGo \citep{sunDirectVoxelGrid2021} is chosen to be our baseline. Due to DirectVoxGo can encode the angular information when projection(color prediction module), and only one angle view is captured at one time in the 2D DSA data, DirectVoxGo is capable of capturing information about contrast flow at different angles/time. We also compared the results of NAF \citep{zha2022naf}. Since NAF uses a different codebase, we have aligned the camera pose and optimization strategy as much as possible, however, numerous hyperparameters could affect training. Moreover, NAF is not applicable to medical reconstruction scenarios involving contrast agent flow. So NAF serves as a rough reference in our study.

\textbf{2D novel view synthesis task}: there are a total of 133 2D DSA images, N of them (N = 4, 10, 20, 30, 40, 50) are selected at approximately equal intervals as training views, and all others are used as test views. Training views are employed for reconstruction, while test views are rendered and subsequently compared to the ground truth for evaluation. We consider the comparison with NAF \citep{zha2022naf} and DirectVoxGo \citep{sunDirectVoxelGrid2021} methods, and give the results of PSNR and SSIM accordingly.

From Fig.~\ref{fig:realresult} and Tab.~\ref{tab:resultoverview}, it can be observed that the predicted results closely match the ground truth. TiAVox demonstrates significant improvements in accuracy and reduces training time (from 21.1 min to 4.9 min on a single V100 GPU) compared to DirectVoxGo. Notably, for a limited number of views (below 30), TiAVox surpasses DirectVoxGo's performance by a substantial margin. In comparison with NAF, which doesn't effectively model the contrast agent flow, NAF tends to produce blurrier vessel reconstructions, especially in scenarios devoid of contrast agents, leading to diminished PSNR values, as evidenced in Fig.~\ref{fig:realresult}. Once the number of reconstructed views reaches 30, the 2D PSNR of TiAVox exceeds 31.2, which is achieved by less than 1/4 views of FDK methods(133 views). It's also salient to note that while segmentation outcomes may vary across different views, TiAVox's predictions maintain consistency.

\begin{figure}[!t]
\begin{center}
\includegraphics[width=\linewidth]{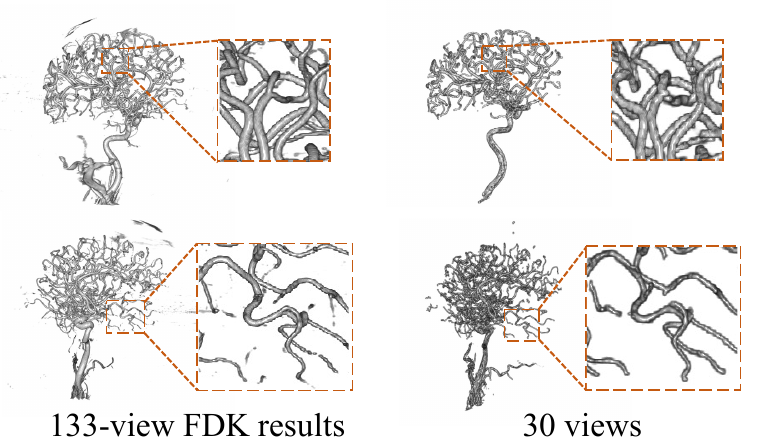}
\end{center}
\caption{3D visualization of 133-view FDK 3D result and TiAVox result of 30 views. TiAVox result is 4D(3D space+1D time), we select a specific time result from TiAVox to compare with 133-view FDK 3D result. Besides, the result obtained from TiAVox primarily represents the contrast agent signal rather than the vasculature itself.}
\label{fig:realresult_large}
\end{figure}

\textbf{3D reconstruction task}: Given that the reconstructed outcomes are not in perfect alignment with the gold standard (133-view FDK 3D result) due to difference in generation method, quantitative evaluation using the PSNR metric is unfeasible. We give the results visualized by slicer software. As depicted in Fig.~\ref{fig:realresult}, our reconstructions are intricate, but they don't match the clarity of the gold standard. The divergence is attributable to the nature of TiAVox results, which fundamentally offer 4D DSA reconstructions, capturing a 3D snapshot at a specific time. In contrast, the 133-view FDK result provides static 3D visuals. Furthermore, the flow of the radiocontrast agent can obscure the boundaries of blood vessels. This is due to the fact that the reconstructed contrast agent signal might not consistently fill the vessels, and the flow pattern of the agent itself is often irregular and lacks smoothness. Potential enhancements are feasible through post-processing and refined pose alignment. As demonstrated in Fig.~\ref{fig:realresult_large}, even with a limited 30-viewpoint perspective, TiAVox delivers comprehensive reconstructions.

\begin{figure*}[htbp]
\centering
\includegraphics[width=520pt]{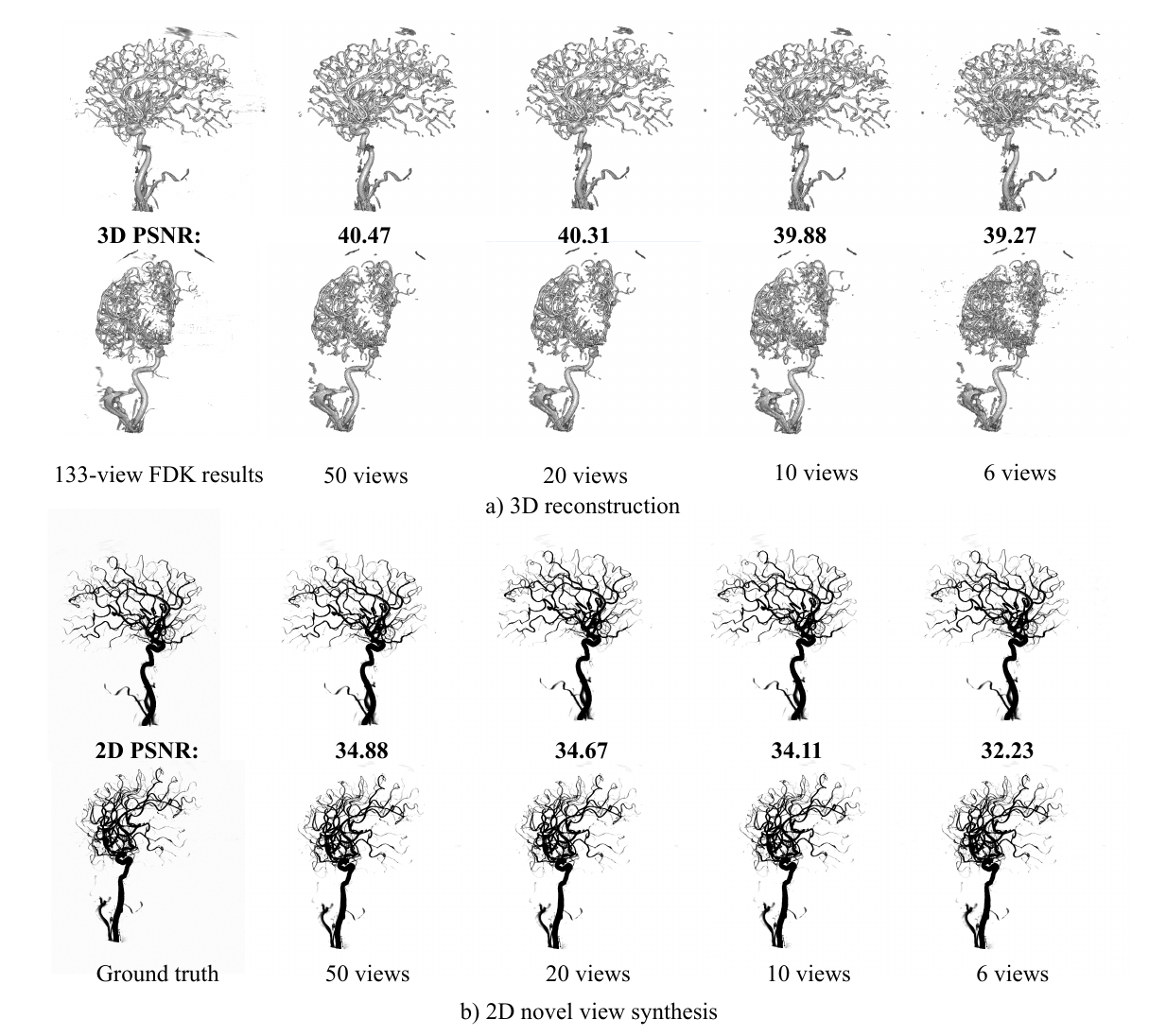}
\caption{Reconstruction result of simulated projections from simulated projections}
\label{fig:idearesult}
\end{figure*}
%
\begin{table*}[htbp]
    \centering
    \renewcommand{\arraystretch}{1.2}
    \caption{2D Novel view synthesis and 3D reconstruction results from simulated images over views (8 patients)}
    \label{tab:idearesultoverview}
    \scalebox{0.9}{\subfloat{\textbf{2D Results of novel view synthesis}}{\label{tab:psnr}}}

    \scalebox{0.9}{
    \begin{tabular}{>{\centering\arraybackslash}p{4.5cm} *{12}{>{\centering\arraybackslash}p{0.6cm}}>{\centering\arraybackslash}p{1.5cm}}
        \toprule
        \multicolumn{1}{c}{Views} & \multicolumn{2}{c}{4 views} & \multicolumn{2}{c}{10 views} & \multicolumn{2}{c}{20 views} & \multicolumn{2}{c}{30 views} & \multicolumn{2}{c}{40 views} & \multicolumn{2}{c}{50 views} & \multicolumn{1}{c}{Training time} \\
         & PSNR$\uparrow$ & SSIM$\uparrow$ &  PSNR$\uparrow$ & SSIM$\uparrow$ &  PSNR$\uparrow$ & SSIM$\uparrow$ & PSNR$\uparrow$ & SSIM$\uparrow$ & PSNR$\uparrow$ & SSIM$\uparrow$ & PSNR$\uparrow$ & SSIM$\uparrow$ & (30 views) \\
        \midrule
        NAF\citep{zha2022naf} & 17.84 &  0.68 &  27.55 &  0.89 &  29.90 &  0.91 &  30.34 &  0.91 &  31.01 &  0.91 &  32.12 &  0.92 & \textgreater 30 min\\
        DirectVoxGo\citep{sunDirectVoxelGrid2021} & 17.33 &  0.79 &  22.62 &  0.88 &  30.53 &  0.92 &  32.29 &  0.92 &  32.98 &  0.92 &  33.45 &  0.92 & 13.2 min\\
        TiAVox(ours) & \textbf{27.47} & \textbf{0.90} & \textbf{34.32} & \textbf{0.93} & \textbf{34.86} & \textbf{0.93} & \textbf{34.98} & \textbf{0.93} & \textbf{35.01} & \textbf{0.93} & \textbf{35.06} & \textbf{0.93} & \textbf{2.8 min}\\

        \bottomrule
    \end{tabular}}

    \vspace{0.2cm}
    \scalebox{0.9}{\subfloat{\textbf{Results of 3D reconstruction}}{\label{tab:ssim}}}\vspace{0.15cm}

    \scalebox{0.9}{
    \begin{tabular}{>{\centering\arraybackslash}p{4.5cm} *{12}{>{\centering\arraybackslash}p{0.6cm}}}
        \toprule
        \multicolumn{1}{c}{Views} & \multicolumn{2}{c}{4 views} & \multicolumn{2}{c}{10 views} & \multicolumn{2}{c}{20 views} & \multicolumn{2}{c}{30 views} & \multicolumn{2}{c}{40 views} & \multicolumn{2}{c}{50 views} \\
         & PSNR$\uparrow$ & SSIM$\uparrow$ &  PSNR$\uparrow$ & SSIM$\uparrow$ &  PSNR$\uparrow$ & SSIM$\uparrow$ & PSNR$\uparrow$ & SSIM$\uparrow$ & PSNR$\uparrow$ & SSIM$\uparrow$ & PSNR$\uparrow$ & SSIM$\uparrow$ \\
        \midrule
        NAF\citep{zha2022naf} & 36.06 &  0.96 &  40.09 &  0.99 &  40.71 &  0.99 &  40.62 &  0.99 &  40.61 &  0.99 &  40.51 &  \textbf{1.00}\\
        TiAVox(ours) & \textbf{38.58} & \textbf{0.99} & \textbf{41.40} & \textbf{1.00} & \textbf{41.99} & \textbf{1.00} & \textbf{42.13} & \textbf{1.00} & \textbf{42.19} & \textbf{1.00} & \textbf{42.19} & \textbf{1.00}\\
        \bottomrule
    \end{tabular}}
\end{table*}

\subsection{Reconstruction from simulated projections}

We show reconstruction results from simulated projections, in which 3D reconstructions can be evaluated by quantitative metrics, and show TiAVox's results without errors in pose parameters, non-vascular noise. We directly reconstruct the 2D image, using 4-50 views(4, 10, 20, 30, 40, 50) for reconstruction, and present the corresponding 3D reconstruction and novel view synthesis results.

\textbf{2D novel view synthesis task}: In our experiments, we maintained consistency with the setup used for clinically collected images. We also compared our results with DirectVoxGo \citep{sunDirectVoxelGrid2021} and NAF \citep{zha2022naf}. As illustrated in Fig~.\ref{fig:idearesult} and Tab~.\ref{tab:idearesultoverview}, with only 10 views, the 2D PSNR surpasses 34. Concurrently, TiAVox displays marked improvements in both accuracy and speed compared to DirectVoxGo and NAF. Notably, TiAVox's reconstruction results using just 10 views are on par with results of TiAVox trained with 50 views. In contrast, DirectVoxGo (DVG) exhibits suboptimal reconstruction performance under sparse view conditions.

\textbf{3D reconstruction task}: We directly compared the reconstructed attenuation coefficients with the 133-view reconstructed FDK results to obtain 3D PSNR as well as 3D SSIM metrics. It is worth noting that 133-view FDK 3D result was not directly utilized during the training process. In our study, a comparison was conducted with NAF. Nevertheless, we did not consider the 3D PSNR/SSIM metrics of DirectVoxGo, because DirectVoxGo's learning of the contrast agent flow through the color prediction module, along with the employment of a distinct rendering formula, results in the density in DirectVoxGo lacking the same physical interpretation as the attenuation coefficients.

\begin{table}[!t]
\centering
\caption{Ablation studies about architecture and training strategy(30 views)}
\label{tab:ablation_study}
\centering
\setlength{\tabcolsep}{1.3mm}
\scalebox{0.8}
{
\begin{tabular}{cccccc}
\toprule
Multiple temporal & Progressive & Space & PSNR & SSIM  & Training \\ 
resolution & temporal resolution & coarse stage & & & time \\
\midrule
\XSolidBrush & \XSolidBrush & \XSolidBrush  & 29.02 & 0.94 & 2.3 min  \\
 \Checkmark & \XSolidBrush & \XSolidBrush  &  32.38 & 0.96 & 4.4 min    \\
 \Checkmark & \Checkmark &  \XSolidBrush &  32.56 & 0.97 & 4.9 min    \\
\Checkmark & \Checkmark & \Checkmark  &  32.48 & 0.97 &  8.6 min    \\
\bottomrule
\end{tabular}}

\end{table}

\begin{table}[tb]
\centering
\renewcommand{\arraystretch}{1.3}
\caption{Ablation studies about space and temporal resolution(30 views)}
\label{tab:ablation_study2}
\centering
\setlength{\tabcolsep}{1.3mm}
\scalebox{0.8}
{

\begin{tabular}{cccccc}
\toprule
Spatial resolution & Temporal resolution  & PSNR & SSIM & Training time\\
\midrule
${220^3}$ & 4 & 32.00 & 0.96 & 3.1 min\\
${320^3}$ & 4 & 32.38 & 0.96 & 4.4 min\\
${420^3}$ & 4 & 32.13 & 0.96  & 6.5 min\\
\midrule
${320^3}$ & 1 & 29.02 & 0.94 & 2.3 min\\
${320^3}$ & 2 & 30.76 & 0.96 & 3.0 min\\
${320^3}$ & 4 & 32.38 & 0.96 & 4.4 min\\
${320^3}$ & 6 & 32.38 & 0.96 & 7.2 min\\
${320^3}$ & 8 & 32.20 & 0.96 & 22.2 min\\

\bottomrule
\end{tabular}}

\end{table}

As can be seen from Fig.~\ref{fig:idearesult} and Tab.~\ref{tab:idearesultoverview}, the result of the 3D reconstruction is almost the same as the label, the details and blood vessel information are perfect, and the PSNR reaches about 42. In contrast to the 133-view FDK 3D results, TiAVox demonstrates the capability to attain great reconstruction results with fewer than 10 views. The metrics and visual results emphasize TiAVox's proficiency in executing the reconstruction task using simulated projections.

\subsection{Ablation studies}
In this section, we performed ablation experiments on each part of the improvement to demonstrate validity. The experiments in this section use clinically collected data from 1 patient and 2D PSNR/SSIM results are reported from novel view synthesis task of 30-view reconstruction. Training time includes time for reconstruction and does not include time for storing the model, it is measured on the V100 GPU.

\textbf{Effects of architecture and training strategy}. For architecture, utilizing multiple temporal resolutions can yield significant performance improvements. For the training strategy, we studied the effects of increasing the temporal resolution of TiAVox during training and maintaining the coarse stage reconstruction in DirectVoxGo. As can be seen from Tab.~\ref{tab:ablation_study}, increasing the temporal resolution of TiAVox from coarse to fine can further improve the results; adding the coarse stage slightly enhances the results, albeit at the expense of increased training time, thus establishing a trade-off between time and accuracy. Unless specified, the method we employed does not include a coarse stage.

\textbf{Effect of TiAVox spatial resolution and temporal resolution}. Excessive spatial and temporal resolution may result in an inflation of learnable parameters. Consequently, achieving improved results would demand a higher number of supervised signals (training views). Conversely, insufficient temporal resolution fails to comprehensively represent the dynamic changes in radiocontrast agent, leading to diminished reconstruction quality. Given the sparse 2D supervision, it becomes imperative to weigh the collective supervisory information and scene dynamics' intricacy in determining the optimal temporal resolution. In our experiments, optimal results emerged when spatial resolution was set to ${320^3}$ and temporal resolution to 4.

\section{Discussion}
In this study, We propose a simple but effective model TiAVox for sparse-view 4D DSA reconstruction, its advantages in: (a) TiAVox can reconstruct 4D/3D/2D result from the real clinical scene of radiocontrast agent flow, even if there is only one 2D DSA image at each moment; (b) TiAVox straightly learns the density (attenuation coefficient) of biological tissues without using a neural network that is difficult to explain; (c) Different from CNN, TiAVox does not require 3D supervision, and only needs 30 or less views to reconstruct. 

\begin{figure}[t]
\centering
\includegraphics[width=250pt]{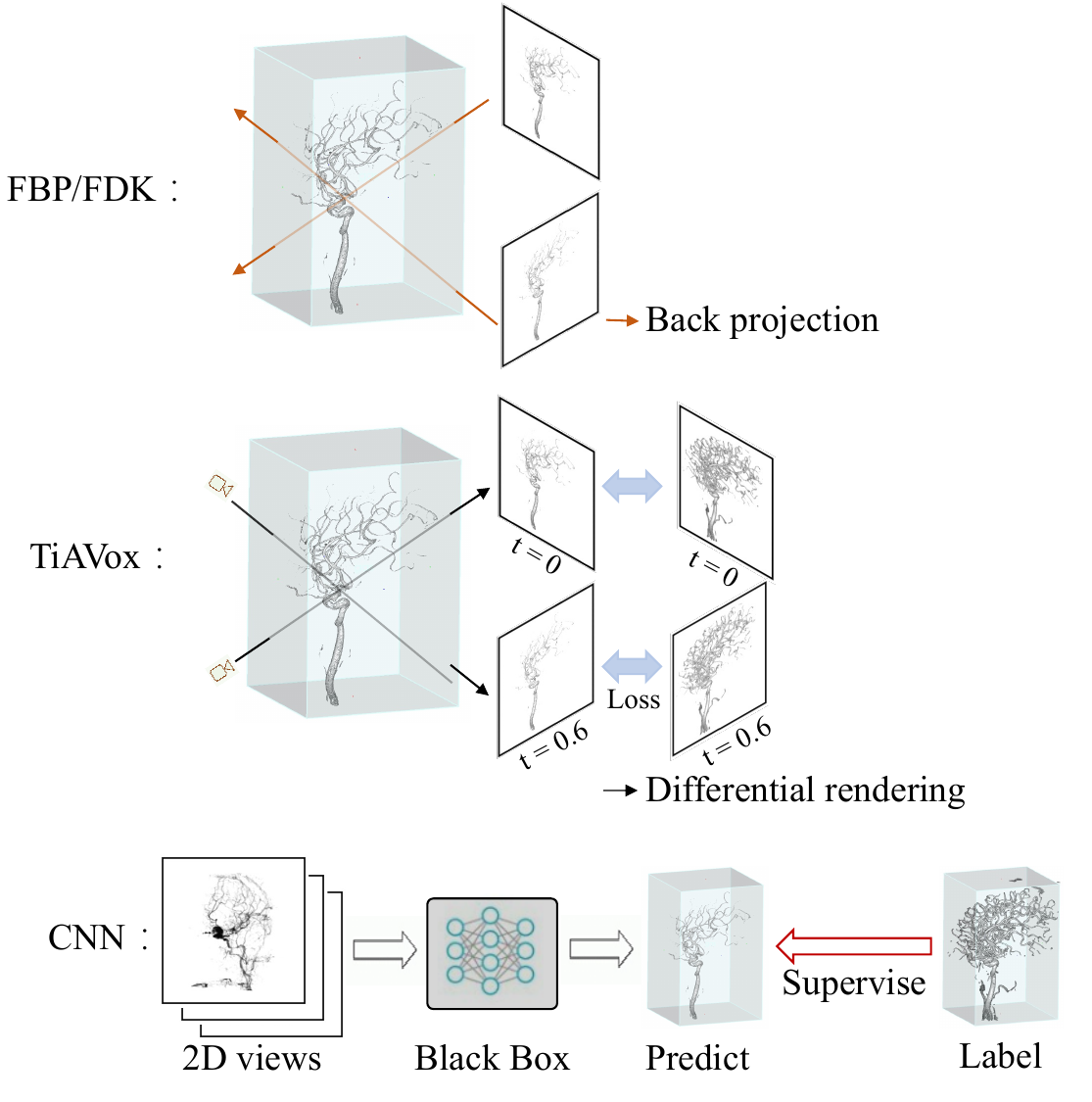}
\caption{Comparsion of FDK, TiAVox and CNN}
\label{fig:3method}
\end{figure}

\textbf{Interpretability analysis}. The FDK algorithm and TiAVox both possess a high degree of interpretability, as depicted in Fig.~\ref{fig:3method}. At the methodological level, the FDK algorithm calculates the voxel density value for each point based on a 2D projection, meaning that an increase in the reconstruction resolution necessitates a greater number of projections. This need becomes particularly problematic in sparse-view reconstructions, where significant artifacts are introduced. Conversely, TiAVox uses an iterative back-propagation algorithm to update voxel values and minimize the projection error, taking into account the multi-view consistency information among different projections. This method significantly reduces the number of required projections. However, CNNs often pose challenges due to their "black-box" nature, particularly in fields such as medicine and autonomous driving. The feature extraction process of CNNs remains largely uninterpretable, making their application in certain domains challenging.

The efficacy of TiAVox in sparse-view reconstruction tasks is demonstrated under both clinically collected and simulated data. Each of these distinct experimental settings presents its unique set of merits and drawbacks. The clinically collected data allows for direct appraisal of the method's clinical applicability, though the quantitative evaluation of the 3D reconstruction outcomes is not feasible due to alignment discrepancies between the gold standard result and our reconstruction results. In contrast, simulation data facilitates a more comprehensive analysis of TiAVox, inclusive of quantification, albeit without incorporating the noise and radiocontrast agent flow inherent in clinical scenarios. Therefore, we exhibit reconstruction outcomes in both situations. It is observable that TiAVox exhibits excellent reconstruction performance with simulated data and also yields remarkable results with clinically collected data.

\section{LIMITATION AND CONCLUSION}
This paper proposes a simple but effective method for sparse-view 4D DSA reconstruction, TiAVox, which also has limitations: 1)	The 3D visualization results reconstructed from clinically collected projections can be improved, for example, to make the blood vessel boundaries clearer. We intend to explore techniques like temporal result fusion, post-processing, and introduce prior knowledge, and so on. 2) Our current approach involves initial segmentation followed by reconstruction. However, segmentation errors can introduce reconstruction noise, further compounded by inaccurate pose information, negatively affecting the results.

Moving forward, our research will prioritize minimizing the view count required for DSA reconstruction, exploring its application across other modalities, interfacing with hardware, and refining pose parameter accuracy, among other considerations.

In summary, our proposed TiAVox enables self-supervised reconstruction of 2D/3D/4D DSA results from clinically collected 2D DSA images using sparse views (less than 30 views), considering radiocontrast agent flow, without any neural network. Such an approach holds promise in reducing radiation exposure in clinical settings, thus diminishing risks to patients. Furthermore, its applicability extends to a range of medical reconstruction contexts, including CT. We have extensively validated our model in both clinic and simulated scenarios.

\bibliographystyle{model1-num-names}
\bibliography{refs.bib}

\end{document}